# Finding the Unicorn: Predicting Early Stage Startup Success through a Hybrid Intelligence Method

*Short Paper*

**Dominik Dellermann**
University of Kassel/Information Systems
Pfannkuchstr.1, 34121 Kassel, Germany
Dellermann@uni-kassel.de

**Nikolaus Lipusch**
University of Kassel/Information Systems
Pfannkuchstr.1, 34121 Kassel, Germany
Nikolaus.lipusch@uni-kassel.de

**Philipp Ebel**
University of Kassel/Information Systems
Pfannkuchstr.1, 34121 Kassel, Germany
Ph.ebel@uni-kassel.de

**Karl Michael Popp**
SAP SE
Dietmar-Hopp-Allee 16, 69190 Walldorf, Germany
karl.michael.popp@sap.com

**Jan Marco Leimeister**
University of Kassel/University of St. Gallen
Information Systems/Institute of Information Management
Pfannkuchstr.1, 34121 Kassel, Germany/Müller-Friedberg-Strasse 8, 9000 St.Gallen, Switzerland
leimeister@uni-kassel.de/janmarco.leimeister@unisg.ch

## Abstract

*Artificial intelligence is an emerging topic and will soon be able to perform decisions better than humans. In more complex and creative contexts such as innovation, however, the question remains whether machines are superior to humans. Machines fail in two kinds of situations: processing and interpreting "soft" information (information that cannot be quantified) and making predictions in "unknowable risk" situations of extreme uncertainty. In such situations, the machine does not have representative information for a certain outcome. Thereby, humans are still the "gold standard" for assessing "soft" signals and make use intuition. To predict the success of startups, we, thus, combine the complementary capabilities of humans and machines in a Hybrid Intelligence method. To reach our aim, we follow a design science research approach to develop a Hybrid Intelligence method that combines the strength of both machine and collective intelligence to demonstrate its utility for predictions under extreme uncertainty.*

# Introduction

Artificial intelligence is an emerging topic and will soon be able to perform administrative decisions faster, better, and at a lower cost than humans. Machines can consistently process large amount of unstructured data to identify pattern and make predictions on future events (e.g. Agrawal and Dhar 2014; Baesens et al. 2016). In more complex and creative contexts such as innovation and entrepreneurship, however, the question remains whether machines are superior to humans. Machines fail in two kinds of situations: processing and interpreting "soft" types of information (information that cannot be quantified) (Petersen 2004) and making predictions in "unknowable risk" situations of extreme uncertainty that require intuitive decision making. In such situations, the machine does not have representative information for a certain outcome and overfits on training data at cost of the live performance of a learner (Attenberg et al. 2015).

One example where both "soft" information signals as well as "unknowable risk" are crucial, is predicting the success of early stage startups. In this case, angel investors, informal investors that devote their private equity in new ventures, face the challenge to decide if the startup at hand is worth financing or not. Angel investors often make decisions before neither the feasibility of a new product nor the existence of a market is proven (Maxwell et al. 2011). In such contexts, angel investors do simply not have enough information to assess the quality of a startup and thus predict the probability of its future success (Dutta and Folta 2016). Moreover, such information might simply not exist under conditions of extreme uncertainty and make the outcome thus unknowable. Consequently, predictions are made for ideas that serve markets, which do not yet exist or novel technologies, where feasibility is still unknown but may provide great returns of investment (Alvarez and Barney 2007). Nevertheless, identifying such unicorns, startups that are highly innovative, disrupt traditional industries, and offer tremendous return is highly relevant.

In these situations, humans are still the "gold standard" for processing "soft" signals that cannot easily be quantified into models such as creativity, innovativeness etc. (Baer and McKool 2014) and make use of an affective judgment tool to recognize pattern in previous decisions: intuition (Huang and Pearce 2015). Using one´s gut feeling proved to be a valuable strategy to deal with extreme uncertainty. However, individual human judges are tainted by bounded rationality in making predictions, which emphasizes that instead of optimizing every decision, humans tend to rely on heuristics (i.e. mental shortcuts) and thus rather focus on highly accessible information (Simon 1955; Kahneman 2011). This often leads to biased interpretation (cognitive processes that involve erroneous assumptions) and may finally result in disastrous predictions (Busenitz and Barney 1997). To solve this problem, research in the field of human computation provides a valuable solution: utilizing the "wisdom of crowds" through collective intelligence (e.g. Brynjolfsson et al. 2016; Larrick et al. 2011; van Bruggen et al. 2010). This is a suitable approach to leverage the benefits of humans in prediction tasks, such as providing subjective evaluation of variables that are difficult to measure objectively through machines (e.g. innovativeness) (Colton and Wiggins 2012) or using their prior domain-specific knowledge to make intuitive decision (Blattberg and Hoch 1990). The aggregation of knowledge and resulting predictions than eliminates the statistical errors of individual human decision makers (Larrick et al. 2011). While each of the methods might work well in separation, weargue that combining the complementary capabilities of humans and machines in a Hybrid Intelligence approach allows to make predictions in contexts of extreme uncertainty such as the case of early startup success through applying formal analysis of "hard" information as well as intuitive decision-making processing also "soft" information

The aim of this research is to develop a method to predict the probability of success of early stage startups. Therefore, we follow a design science research approach (Hevner 2007; Gregor and Hevner 2013) to develop a Hybrid Intelligence method that combines the strength of both machine intelligence such as machine learning techniques to access, process, and structure large amount of information as well as collective intelligence, which uses the intuition and creative potential of individuals while reducing systematic errors through statistical averaging in an ensemble approach (Shmueli and Koppius 2011). We, thus, intend to show that a hybrid approach improves predictions for the success of startups under extreme uncertainty compared to machine or human only methods. As we proceed our research we will empirically test our proposed method to provide and validate a practical solution for predicting firm success under conditions of extreme uncertainty.

Within the scope of this paper, we first developed a taxonomy of signals that are potential predictors for the success of early stage startups based on previous work and domain knowledge (Shmueli and Koppius 2011).

We then designed a method that uses these predictors as input for both machine learning algorithms as well as collective intelligence to individually assess the probability of success and then weights and aggregates the results to a combined prediction outcome. Moreover, we provide an outlook on the next steps of our research project.

This work will thus contribute to several important streams of IS and management research. First, we provide a taxonomy of potential predictors that can be generalized for modelling startup success predictions (e.g. Böhm et al. 2017). Second, this research adds to literature on predictive research in IS and data analytics (e.g. Chen et al. 2012) by introducing a new method for predicting uncertain outcomes under limited information and unknowable risk by combining collective and machine intelligence in a Hybrid Intelligence Method. This approach allows to complement formal analysis of "hard" information and intuitive predictions based on "soft" information. Consequently, our research will offer prescriptive knowledge in this vein (Gregor and Jones 2007). Third, we contribute to previous work on collective intelligence (e.g. Malone et al. 2009; Wooley et al. 2010) by proposing novel applications of the crowd. Finally, we will provide a practical solution that will offer angel investors a useful way to support their financing decisions.

# Related Work

## *Defining the Context: Predicting Startup Success under Extreme Uncertainty*

One way towards understanding predictions in uncertain situations is to examine the mental processes that underlies the cognitive decision-making process. A theory that is particularly helpful in this context is the dual process theory of decision making. The underlying assumption of this theory is that people make use of two cognitive modes, one is characterized by intuition (*system 1*) and one by deliberate analytical predictions (*system 2*) (Tversky and Kahneman 1983; Kahneman 2011).

Predicting the success of early stage ventures is extremely complex and uncertain because frequently just vague ideas are prevalent, prototypes do not yet exist and thus the proof of concept is still pending. Moreover, such ideas might even not have a market yet, but offer great potential of growth in the future (Alvarez and Barney 2007). Consequently, the decision-making context is highly uncertain as neither possible outcomes nor the probability of such are known. This fact can be explained through two concepts: information asymmetry and unknowable risk (Alvarez and Barney 2007; Huang and Pearce 2015).

Information asymmetry describes situation, in which forecasters have incomplete information to decide (Spence 1974). When perfect information is absent, decision makers tend to search for various indicators that signal the likeliness of future outcomes (Morris 1987). In our context, such signals include both "hard" signals that can be easily quantified and categorized (e.g. industry, technology, team size) as well as "soft" signals (e.g. innovativeness, personality of entrepreneur). Humans then try to apply formal analysis to gather signals that support them in making deliberate, rule-based *system 2* decisions (Kahneman 2011). On the other hand, unknowable risk defines situations in which a decision maker cannot gather information that signal a potential outcome or make decisions based on formal analysis because the simply not exist. This may be best compared to the error term of a statistical Bayesian model. Unknowable risk covers unexpected events that describe a deviation from status quo (Kaplan and March 1988). In our context, this means for instance identifying a unicorn startup that gains enormous return that only few would have expected. Formal analytics are not working in these contexts, as representative cases might be missing in previous experience. In such situations, where humans "don't know what they don't know", decision making is mainly based on intuition (*system 1*) rather than formal analysis (Tversky and Kahneman 1983; Huang and Pearce 2015). Thus, predicting the success of early stage startups is a challenging task and the costs of misclassification are high as they might lead to disastrous funding decisions or missing valuable chances for return (Attenberg et al. 2015). Previous research in the context of early stage ventures provides strong evidence the best performance in terms of accuracy are provided by combining both types of predictions: analytical (*system 2*) and intuitive (*system 1*) (Huang and Pearce 2015).

## *Machine Intelligence as a Possible Solution for Predicting Startup Success*

For analytically predicting future events, computational statistical methods become particularly valuable due to progresses in machine learning and artificial intelligence. They can identify, extract and process various forms of data from different sources (e.g. Böhm et al. 2017; Carneiro et al 2017). Statistical modelling allows to make highly accurate and consistent predictions in the context of financing decisions (Yuan et al. 2016), financial return (Craemer et al. 2016), or bankruptcy of firms (Olson et al. 2012) by identifying patterns in the prior distribution of data and thus predict future events. Machine intelligence is, thus, particularly valuable as biases or limited capacity of human decision makers does not taint it. Statistical models are unbiased, free of social or affective contingence, consistently integrate empirical evidence and weigh them optimally and they are not constraint by cognitive resource limitations (Blattberg and Hoch 1990). Consequently, machine intelligence is a suitable approach for making statistical inference based on prior data and they can learn as the data input grows (Jordan and Mitchell 2015). While such machine intelligence approaches are superior in analytically predicting uncertain outcomes by minimizing the problems of information asymmetries and bounded rationality based on prior distributions of "hard" objective variables (e.g. firm age, team size), they are neither able to explain the remaining random error term of such distributions, which we conceptualized through unknowable risk (Aldrich 1999) nor the "soft" and subjective signals of new ventures such as innovativeness, the vision or the fit of the team, or the overall consistency of a new venture (Petersen 2004). Both limitations of machines might lead to costly misclassifications (Attenberg et al. 2015). While progresses in the field of artificial intelligence provide evidence for the applicability of machines in making subjectivity decisions (e.g. Cowgill 2017) intuitive predictions remain the advantage of humans and require the completion of machine capabilities.

### *Towards Complementary Capabilities: Machine and Collective Intelligence*

In this vein, the benefits of human decision makers come into play. Humans are still the "gold standard" for assessing "soft" signals that cannot easily be quantified into models such as creativity and innovativeness (Baer and McKool 2014). Humans are talented at making intuitive predictions by providing subjective judgement of information that is difficult to measure objectively through statistical models (e.g. Einhorn 1974). Moreover, human decision makers can have highly organized domain knowledge that enables them to recognize and interpret very rare information. Such information might lead to outcomes that are difficult to predict and would rather represent outliers in a statistical model (Blattberg and Hoch 1990). Consequently, using human intuition proved to be a valuable strategy for anticipating startup success at an early stage (Huang and Pearce 2015).

However, individual decision makers make errors due to their bounded rationality (Simon 1982; Kahneman 2003). This assumption considers the capacity of the human mind for solving complex problems as rather constraint. Instead of optimizing every decision, individuals tend to engage in limited information accessing to reduce cognitive effort (Hoenig and Henkel 2015). Consequently, they use cognitive heuristics (i.e. mental shortcuts) and simplifying knowledge structures for reducing information-processing demands. One is for example drawing conclusions from a small amount of information or using easy accessible signals (Tversky and Kahneman 1974). Moreover, humans have several biases (cognitive processes that involve erroneous assumptions) that guide the interpretation of information to make predictions (Busenitz and Barney 1997).

Research on human computation provides a solution for these problems (Quinn and Bederson 2011). Collective intelligence, leverages the "wisdom of crowds" to aggregate the evaluations of a large group of humans, thereby, reducing the noise and biases of individual predictions (Atanasov et al. 2017; Cowgill and Zitzewitz 2015; Blohm et al. 2016). The value of crowds compared to individuals underlies two basic principles: error reduction and knowledge aggregation (Larrick et al. 2011; Mellers et al. 2015). Error reduction is because although individual decision makers might be prone to biases and errors, the principle of statistical aggregation minimizes such errors by combining multiple perspectives (Armstrong 2001). Second, knowledge aggregation describes the diversity of knowledge that can be aggregated by combining the experience of multiple decision makers. Such knowledge aggregation enables to capture a fuller understanding of a certain context (Keuschnigg and Ganser 2017). Thus, collective intelligence can assess the probability of uncertain outcomes by accessing more diverse signals and reduce the threat of biased interpretation.

Using collective intelligence enables to complement a machine model by assessing unknowable risk, which cannot be explained through prior distribution but rather from the combined intuition of humans. In particular, machine learning provides advantages in making analytical predictions based on "hard" information while collective intelligence offers benefits in making intuitive predictions taking also "soft" information into account (Figure 1). Previous work, emphasizes the complementary nature of humans and statistical models in making predictions about future events in various settings such as sports, politics, economy, or medicine (Blattberg and Hoch 1990; Ægisdóttir et al. 2006; Einhorn 1972; Meehl 1954; Nagar and Malone 2011). We thus argue that a Hybrid Intelligence Method that combines the complementary capabilities of analytical and intuitive predictions is most accurate for predicting the success of early stage startups.

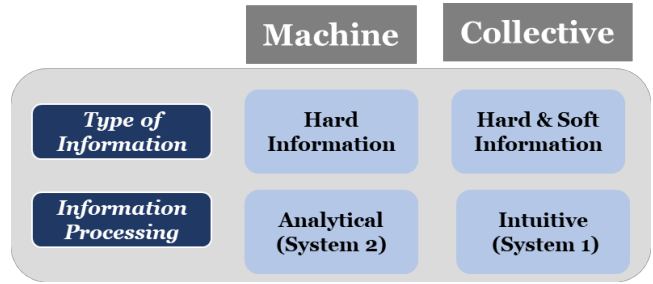

**Figure 1. The Complementary Capabilities of Human and Machine Intelligence**

## Methodology

To develop a method that capitalizes the benefits of both machine and collective intelligence, we followed a design science research (DSR) approach (Hevner et al. 2004; Gregor and Hevner 2013) guided by schematic steps of Shmueli and Koppius (2011). The result of this design science project constitutes a hybrid intelligence method as a new and innovative artifact that helps to solve a real-world problem. Following Hevner's (2007) three cycle view of relevance and rigor we combine inputs from the practical problem domain (relevance) with the existing body of knowledge (rigor). Abstract theoretical knowledge thus has a dual role. First, it addresses the suggestions for a potential solution. Second, the abstract learning from our design serves as blueprint to provide prescriptive knowledge for solving similar problems in the future (Gregor and Jones 2007).

## Development of a Solution

As outlined above, both collective and machine intelligence approaches have specific benefits to predict uncertainty future events. To develop a novel method to predict the future success of early stage startups, we decided to combine an ensemble of machine and collective intelligence due to several reasons. First, combining multi methods has a long tradition in research on forecasting as combined methods and sources should be at least similar or better than the best individual prediction method (Armstrong 2001). Second, previous research considered the strengths of human decision makers and statistical models and anticipated their complementary capabilities in making predictions (Blattberg and Hoch 1990; Nagar and Malone 2011). Third, we argue that for contexts of extreme uncertainty due to both information asymmetry and unknowable risk, a hybrid intelligence method consisting of collective and machine intelligence, is superior to one or the other approach alone. A hybrid intelligence method for predicting the success of new ventures enables to statistically model a prediction by consistently accessing and processing "hard" signals based on Bayesian inference, while collective intelligence allows to access more diverse "soft" signals through aggregation and reducing the systematic errors of individual humans, which also helps to reduce the random error term of predictions that reflects unknowable risk through the benefits of human intuition.

Figure 2 shows our hybrid intelligence method to predict the success of early stage startups. A fully automated system with a large flow of information will require data mining trough web crawling approaches and preprocessing the raw data. Within the scope of this paper we focus on the main part of the hybrid intelligence method by explaining input metrics, the automation process and expert weighting to predict the success of early stage startups.

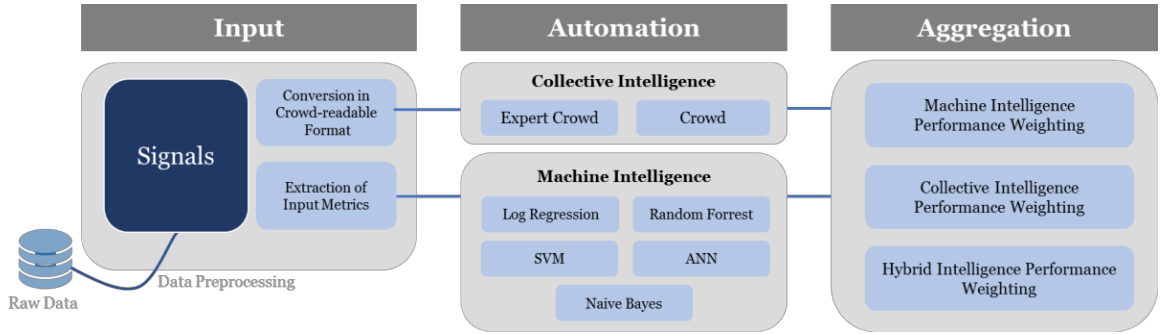

**Figure 2. Our Proposed Hybrid Intelligence Method**

## Objectives of a Solution: Predicting Success of Early Stage Startups

The objective of our solution is to accurately predict the success of early stage startups and thus make accurate predictions under extreme uncertainty. To define the success of early stage startups, we assessed whether they received series A funding and used this funding as a proxy. This is the commonly accepted indicator for success in this context (e.g. Baum and Silverman 2004; Spiegel et al. 2015). Series A funding defines the first venture capital backed funding that allows angel investors to exit the startup. The objective is thus to provide a probabilistic classification of the binary categorical variable series A funding.

## Input: Taxonomy of Signals

| Category | Signals | Reference Examples | Hard Signal | Soft Signal | Machine Input Metric | Crowd-readable Format |
|---|---|---|---|---|---|---|
| Meta | B2B vs B2C | Böhm et al. (2017) | | | • Categorical (Type) | Textual |
| | Industry | Hoenig and Henkel (2015) | | | • Categorical (Type of industry) | Textual |
| | Firm Age | Zahra et al. (2003) | | | • Numeric (# of years) | Numeric |
| | Business Model DNA | Böhm et al. (2017) | | | • Categorical (Cluster=Freemium Platforms/Experience Crowd Users/Long Tail Subscribers/Affiliate Markets/Mass Customizing Orchestrators/Innovative Platforms/E-Commercers/Add-On Layers/Crowdsourcing Platforms/Customized Layers/Hidden Revenue Markets) | Textual |
| Value Proposition | Product Innovativeness | Li (2001); Nambisan (2016) | | | | Graphic and Textual |
| | Technological Hype | Maxwell et al. (2011) | | | • Categorical (Phase in Gartner Hypecycle) | Graphic and Textual |
| | Proof of Concept | Maxwell et al. (2011); | | | • Dummy (Existence of prototype=yes/no) | Textual |
| | Scalability | Huang et al. (2017); Nambisan (2016) | | | | Textual |
| Market | Competition | Landström (1998); Shi et al. (2016) | | | • Numeric (# of competitors)<br>• Numeric (MEAN proximity to competitor based on Shi et al. (2016)) | Graphic and Textual |
| | Revenue Model | Kirsch et al. (2009); Afuah and Tucci (2001); Böhm et al. (2017) | | | • Categorical (Type of revenue model=commission-based/fee-for-service/advertising/subscription/referral/production/mark-up based/other) | Textual |
| Resources | Capital Raised | Baum and Silverman (2004) | | | • Numeric (Amount of capital raised in USD) | Numeric |
| | Team Size | Mason and Stark (2002); Kirsch et al. (2009) | | | • Numeric (# of team members) | Numeric |
| | Team Constellation | Kirsch et al. (2009) | | | • Numeric (# of field background) | Textual |
| | Entrepreneurial Experience | Franke et al. (2006); Maxwell et al. (2011) | | | • Dummy (Previous founded ventures=yes/no) | Textual |
| | Entrepreneurial Vision | Sudek (2006); Huang and Pearce (2015) | | | | Textual |
| | Entrepreneurial Education | Baum and Silverman (2004); Franke et al. (2006); Maxwell et al. (2011) | | | • Categorical (Level of education=high-school, bachelor, master, PhD) | Graphic and Textual |
| Commitment of 3. Party Support | Knowledge Support | Maxwell et al. (2011); Song et al. (2008); Zahra and Bogner (2000) | | | • Categorical (Type of support=incubator/accelerator/business angel/university) | Textual |
| | Financial Support | Baum and Silverman (2004) | | | • Categorical (Type of funding=state support/equity backed/bank financing) | Textual |
| | Proof of Value | Shepherd and Zacharakis (1999); Mason and Stark (2002) | | | • Numeric (# of pilot customers) | Numeric and Textual |
| | Web Analytics | Silver (2012) | | | • Website visits/average duration/backlinks/bounce rate | Numeric and Textual |
| | Social Media Analytics | Silver (2012) | | | • Twitter follower/# of tweets/sentiment of tweets | |

**Table 1. Taxonomy of Signals for Prediction Input**

To identify relevant metrics for the input of the hybrid intelligence automation we first conducted a literature analysis (Webster and Watson 2002) and then conducted interviews (n=15; average duration 60 min) and focus group workshops (n=6; 3-7 participants; average duration 90 min) with experienced angel investors to iteratively combine the findings from the literature review with practical relevant factors following the procedure of Nickerson et a. (2013). Based on these findings, we developed a taxonomy of relevant input metrics for a hybrid intelligence method in the context of predicting the success of early stage ventures (e.g. Song et al. 2008) (see Table 1). As we focus on early stage startups, we do not include financial metrics, which are barely relevant in this stage (e.g. Maxwell et al. 2011).

## Machine Input Metrics

From this taxonomy, we choose the "hard" signals that can be quantified and translated into machine readable format. We, therefore, used the metrics as displayed in Table 1.

## Conversion in Crowd-readable Format

To automate the prediction through collective intelligence, the signals need to be converted into crowd-readable format. Therefore, we designed a GUI (graphical user interface) that consist of all "hard" and "soft" signals of our taxonomy (Table 1). This ensures that each human can make a prediction based on a comprehensive representation of the startup. A graphical approach was chosen to improve the information extraction capabilities that are necessary for making these predictions (e.g. Todd and Benbasat 1999). The GUI is based on a standardized ontology that was developed from our taxonomy of signals (Burton-Jones and Weber 2014).

### *Automation*

## Machine Intelligence

To apply machine intelligence, we use the above-mentioned metrics as input for probabilistic classification of series A funding. We therefore choose the following learning algorithms due to their different methodological benefits:

1. *Logistic Regression*: a very well-known linear regression algorithm used as the baseline algorithm. Frequently applied for binary choice models (Magder and Hughes 1997).

2. *Naives Bayes*: Bayesian parameter estimation problem based on some known prior distribution (Valles and Varas 2012).

3. *Support Vector Machine (SVM):* classification algorithm based on a linear discriminant function, which uses kernels to find a hyperplane that separates the data into different classes (Vapnik 1998).

4. *Artificial Neural Network (ANN):* a biologically inspired, non-parametric learning algorithm that can model extremely complex non-linear function (Haykin 2009).

5. *Random Forests:* a popular ensemble method that minimizes variance without increasing bias by bagging and randomizing input variables (Breiman 2001).

## Collective Intelligence

The collective intelligence automation is designed as judgment task (Riedl et al. 2013). This asks each human participant of the crowd to rate the startup on a multi-dimensional 7-point Likert scale, which provides the most accurate results in collective intelligence prediction (Blohm et al. 2016). Each startup is then assessed along the most relevant dimensions to indicate success: feasibility, scalability, and desirability (Fitzsimmons and Douglas 2011). To execute the actual rating process, we use two instances of collective intelligence: a crowd of non-experts and an expert crowd. This is consistent with previous research that indicates that a larger group of non-experts as well as a smaller group of domain experts are both capable to predict uncertain outcomes (Klein and Garcia 2015; Mannes et al. 2014).

1. *Crowd*: open call to a crowd of non-experts. 16-20 humans judge each startup to predict its success and leverage the benefits of collective intelligence in non-expert samples (Keuschnigg and Ganser 2017). The non-expert crowd is requested through a crowd working platform (Amazon Mechanical Turk). Language proficiency is ensured through choosing US residents as our startup descriptions were generated in English and quality selection is conducted through a minimum HIT approval rate of 95% as selection criteria. Each participant is financially compensated.

2. *Expert crowd:* restricted open call to humans that have high expertise and domain knowledge (i.e. business angels, startup mentors, consultants) 5-7 experts judge each startup to predict its success and leverage the benefits of collective intelligence in expert samples (Mannes et al. 2014). The expert crowd is requested through a startup mentor network and will not be not financially compensated.

The collective intelligence automation is then calculated as unweighted average that combines each crowd member´s ordinal evaluations into a single score indicating the probabilistic classification of series A funding success, which provides the most accurate combined predictions (Keuschnigg and Ganser 2017).

### Aggregation

To benefit from the hybrid intelligence method, we finally need to aggregate the predictions of each individual human and machine intelligence approach to generate the output of probability of series A funding. We thus will apply two types of aggregation. First, we use simple unweighted averaging as baseline approach which proved to be accurate for many types of predictions (Armstrong 2001; Keuschnigg and Ganser 2017). Second, we follow the idea of Craemer et al. (2016) of combining different individual predictions of the crowd and machine learning algorithms according to their performance (Archak et al. 2011). For this purpose, we propose three of this weighting algorithms to identify the most accurate one. Machine intelligence performance weighting that is based on the predictions of only machine learning algorithms. Collective intelligence performance weighting that is based on the prediction of the crowd. We argue that in contexts involving consumer product startups, non-expert crowds might be superior as they represent the voice of potential customers and providing demand side knwoeldge whereas expert crowds will be more accurate in B2B settings (e.g. Magnusson et al. 2016). Finally, hybrid intelligence performance weighting combines the predictions of machines and humans to predict the success outcome.

## Planned Implementation and Evaluation

To evaluate our method, we will implement the input, automation, and aggregation procedures and evaluate their performance. We started by creating a data set of startups by combining information on the signals in our taxonomy from several databases (Crunchbase, Mattermark, and Dealroom). Thereby, we focused on tech startups as the high-tech context is particularly tainted with extreme uncertainty. The data sample consists of 1500 startups from different industries and all extracted signals from the taxonomy. Part of the sample received series A funding while the other part did not. All startups in our dataset are labeled accordingly. We preprocessed the "hard" signals for the machine algorithms and converted all "hard" and "soft" signals into crowd-readable format (an ontology based GUI). In the next steps, we will standardize the input metrics and test to avoid overfitting (Carneiro et al 2017). The automation process of the "hard" and "soft" signals will then be conducted as described in the previous section.

The dataset will then be randomly split into two sub-sets, one for training and one for testing. The training data is used to create the prediction models. For our purpose, we choose a 10-fold cross validation approach to split our data set into ten mutually exclusive sub-sets of approximately equal size. The idea behind 10-fold cross validation is to minimize the bias associated with the random sampling of the training and holdout data samples. Each of our proposed prediction approaches (i.e. machine algorithms and the crowd) is then trained and tested ten times with the same ten folds, which means the algorithm is trained on nine folds and tested on the remaining single fold. Cross validation then estimates the overall accuracy of an algorithm by calculating the mean accuracy (e.g. Olson et al. 2012).

The evaluate the performance of our method we will use the Matthews correlation coefficient (MCC) that is a well-known balanced performance measure for binary classification, when the classes within the data are of very different sizes (Matthews 1975). We choose this measure as our data set is biased towards successful startups although we also identified a large amount of failed ventures, which is a common limitation of such databases (e.g. Böhm et al. 2017). We will then use logarithmic regression as baseline algorithm (e.g. Craemer et al. 2016) and compare the performance of each individual machine intelligence algorithm, crowd prediction, and weighting algorithm through a two-way analysis of variance (ANOVA) (Bradley 1997). We therefore intend to identify the best combined approach and aim at showing that Hybrid Intelligence approach provides superior results than a machine or human only prediction.

## Conclusion

Predicting the success of early stage startups is a challenging task and the costs of misclassification is high as it might lead to disastrous funding decisions are missing valuable chances for return. To make predictions in such contexts, we propose to combine the complementary capabilities of machine and human intelligence. While machines are particularly beneficial in consistently processing large amount of "hard" signals that indicate the success of a new venture humans are superior in interpreting "soft" signals such as the personality of an entrepreneur or the innovativeness of a new product. Moreover, humans can leverage their intuition to identify valuable startups that cannot be found by relying on previous data. To overcome the constraints of bounded rationality of individuals, we thus suggest leveraging collective intelligence. To reach our aim, we developed a preliminary Hybrid Intelligence method that we will initially evaluate as we proceed our research. In the next steps, we will then also test its applicability for other outcome variables in the context of startups (e.g. growth, survival rate etc.) and other contexts of extreme uncertainty (e.g. innovation in general). Moreover, we intend to assess the relevance of accuracy and transparency with potential users of this method and if they are more willing to take advice when human sources are included (e.g. Önkal et al. 2009). We expect our research to make several contributions to both academia and practice. First, we provide a taxonomy of potential predictors that can be generalized for modelling startup success predictions (e.g. Böhm et al. 2017). Second, this research adds to literature on predictive research in IS and data analytics (e.g. Chen et al. 2012) by introducing a new method for predicting uncertain outcomes under limited information and unknowable risk by combining collective and machine intelligence in a Hybrid Intelligence method. This approach allows to complement formal analysis of "hard" information and intuitive predictions based on "soft" information. Such hybrid method might be valuable for other settings of extreme uncertainty as well. Consequently, our research will offer prescriptive knowledge in this vein that might be generalizable for data science methods in general (Gregor and Jones 2007). Third, we contribute to previous work on collective intelligence (e.g. Malone et al. 2009; Wooley et al. 2010) by proposing novel applications of machines and crowd. We argue that our proposed approach can augment the capabilities of collective intelligence in general. While we use a parallel approach in this paper, further research might explore how machine intelligence might be leveraged as feedback for the crowd and thus point towards more collaborative interactive approaches (e.g. Calma et al. 2016). Finally, we provide a useful solution for a practical prediction problem that may support angel investors in making decisions and potentially reduce the frequency of bad investment decisions.

## Acknowledgments

This paper presents research that was conducted in context of the project "CrowdServ" (funding number: 02K14A210, managed by the Project Management Agency Karlsruhe - PTKA) and funded by the German Federal Ministry of Education and Research (BMBF) (www.crowdserv.de). The responsibility for the content of this publication remains solely with the authors.

## References

Afuah, A., and Tucci, C. L. 2003. "A model of the Internet as creative destroyer," *IEEE Transactions on Engineering Management* (50:4), pp. 395–402.

Agarwal, R., and Dhar, V. 2014. "*Editorial—Big data, data science, and analytics: The opportunity and challenge for IS research*" Information Systems Research (25:3), pp. 443-448.

Aldrich, H. 1999. *Organizations evolving:* Sage.

Alvarez, S. A., and Barney, J. B. 2007. "Discovery and creation: Alternative theories of entrepreneurial action," *Strategic Entrepreneurship Journal* (1:1-2), pp. 11–26.

Archak, N., Ghose, A., and Ipeirotis, P. G. 2011. "Deriving the pricing power of product features by mining consumer reviews," *Management Science* (57:8), pp. 1485–1509.

Armstrong, J. S. 2001. *Principles of forecasting: a handbook for researchers and practitioners:* Springer Science & Business Media.

Atanasov, P. D., Rescober, P., Stone, E., Swift, S. A., Servan-Schreiber, E., Tetlock, P. E., Ungar, L., and Mellers, B. 2017. "Distilling the wisdom of crowds: Prediction markets versus prediction polls," *Management science* (63:3), pp. 691–706.

Attenberg, J., Ipeirotis, P., and Provost, F. 2015. "Beat the Machine: Challenging Humans to Find a Predictive Model's "Unknown Unknowns"," *Journal of Data and Information Quality (JDIQ)* (6:1), pp. 1-15.

Baer, J., and McKool, S. S. 2014. "The gold standard for assessing creativity," *International Journal of Quality Assurance in Engineering and Technology Education (IJQAETE)* (3:1), pp. 81–93.

Baesens, B., Bapna, R., Marsden, J. R., Vanthienen, J., and Zhao, J. L. 2016. "Transformational issues of big data and analytics in networked business," *MIS Quarterly* (40:4), pp. 807–818.

Baum, J. A. C., and Silverman, B. S. 2004. "Picking winners or building them? Alliance, intellectual, and human capital as selection criteria in venture financing and performance of biotechnology startups," *Journal of Business Venturing* (19:3), pp. 411–436.

Baum, J. A. C., and Silverman, B. S. 2004. "Picking winners or building them? Alliance, intellectual, and human capital as selection criteria in venture financing and performance of biotechnology startups," *Journal of Business Venturing* (19:3), pp. 411–436.

Blattberg, R. C., and Hoch, S. J. 1990. "Database models and managerial intuition: 50% model+ 50% manager," *Management Science* (36:8), pp. 887–899.

Blohm, I., Riedl, C., Füller, J., and Leimeister, J. M. 2016. "Rate or trade? identifying winning ideas in open idea sourcing," *Information Systems Research* (27:1), pp. 27–48.

Böhm, M., Weking, J., Fortunat, F., Müller, S., Welpe, I., and Krcmar, H. 2017. "The Business Model DNA: Towards an Approach for Predicting Business Model Success," *Proceedings of the WI 2017.*

Bradley, A. P. 1997. "The use of the area under the ROC curve in the evaluation of machine learning algorithms," *Pattern recognition* (30:7), pp. 1145–1159.

Breiman, L. 2001. "Random forests," *Machine learning* (45:1), pp. 5–32.

Brynjolfsson, E., Geva, T., and Reichman, S. 2016. "Crowd-squared: amplifying the predictive power of search trend data," *MIS Quarterly* (40:4), pp. 941–961.

Burton-Jones, A., and Weber, R. 2014. "Building conceptual modeling on the foundation of ontology," *Computing Handbook, Third Edition: Information Systems and Information Technology*. Chapman and Hall.

Busenitz, L. W., and Barney, J. B. 1997. "Differences between entrepreneurs and managers in large organizations: Biases and heuristics in strategic decision-making," *Journal of Business Venturing* (12:1), pp. 9–30.

Carneiro, N., Figueira, G., and Costa, M. 2017. "A data mining based system for credit-card fraud detection in e-tail," *Decision Support Systems* (96:3), pp. 91–101.

Chen, H., Chiang, R. H. L., and Storey, V. C. 2012. "Business intelligence and analytics: From big data to big impact," *MIS Quarterly* (36:4), pp. 1165–1188.

Colton, S., and Wiggins, G. A. (eds.) 2012. *Computational creativity: The final frontier?* IOS Press.

Cowgill, B. 2017. *Automating Judgement and Decision-making: Theory and Evidence from Résumé Screening.* http://www.sole-jole.org/17446.pdf. Accessed 6 September 2017.

Cowgill, B., and Zitzewitz, E. 2015. "Corporate prediction markets: Evidence from google, ford, and firm x," *The Review of Economic Studies* (82:4), pp. 1309–1341.

Creamer, G. G., Ren, Y., Sakamoto, Y., and Nickerson, J. V. 2016. "A Textual Analysis Algorithm for the Equity Market: The European Case," *The Journal of Investing* (25:3), pp. 105–116.

Dutta, S., and Folta, T. B. 2016. "A comparison of the effect of angels and venture capitalists on innovation and value creation," *Journal of Business Venturing* (31:1), pp. 39–54.

Ægisdóttir, S., White, M. J., Spengler, P. M., Maugherman, A. S., Anderson, L. A., Cook, R. S., Nichols, C. N., Lampropoulos, G. K., Walker, B. S., and Cohen, G. 2006. "The meta-analysis of clinical versus statistical prediction project: Fifty-six years of accumulated research on clinical versus statistical prediction," *The Counseling Psychologist (34:3),* pp. 341–382.

Einhorn, H. J. 1972. "Expert measurement and mechanical combination," *Organizational Behavior and Human Performance* (7:1), pp. 86–106.

Franke, N., Gruber, M., Harhoff, D., and Henkel, J. 2006. "What you are is what you like—similarity biases in venture capitalists' evaluations of start-up teams," *Journal of Business Venturing* (21:6), pp. 802–826.

Gregor, S., and Hevner, A. R. 2013. "Positioning and presenting design science research for maximum impact," *MIS Quarterly* (37:2), pp. 337–355.

Gregor, S., and Jones, D. 2007. "The anatomy of a design theory," *Journal of the Association for Information Systems* (8:5), p. 312.

Haykin, S. S. 2009. *Neural networks and learning machines:* Pearson Upper Saddle River, NJ, USA.

Hevner, A. R. 2007. "A three-cycle view of design science research," *Scandinavian Journal of Information Systems* (19:2), p. 4.

Hevner, A. R., March, S. T., Park, J., and Ram, S. 2004. "Design science in information systems research," *MIS Quarterly* (28:1), pp. 75–105.

Hoenig, D., and Henkel, J. 2015. "Quality signals? The role of patents, alliances, and team experience in venture capital financing," *Research Policy* (44:5), pp. 1049–1064.

Huang, J., Henfridsson, O., Liu, M. J., and Newell, S. 2017. "Growing on steroids: rapidly scaling the user base of digital ventures through digital innovation," *MIS Quarterly* (41:1), pp. 301–314.

Huang, L., and Pearce, J. L. 2015. "Managing the unknowable: The effectiveness of early-stage investor gut feel in entrepreneurial investment decisions," *Administrative Science Quarterly* (60:4), pp. 634–670.

Jordan, M. I., and Mitchell, T. M. 2015. "Machine learning: Trends, perspectives, and prospects," *Science* (349:6245), pp. 255–260.

Kahneman, D. 2003. "A perspective on judgment and choice: mapping bounded rationality," *American Psychologist* (58:9), p. 697.

Kahneman, D. 2011. *Thinking, fast and slow:* Macmillan.

Kahneman, D., and Tversky, A. 1982. "On the study of statistical intuitions," *Cognition* (11:2), pp. 123–141.

Kamar, E. 2016. "Directions in hybrid intelligence: complementing AI systems with human intelligence," *Early Career Track*.

Keuschnigg, M., and Ganser, C. 2017. "Crowd Wisdom Relies on Agents' Ability in Small Groups with a Voting Aggregation Rule," *Management Science* (63:3), pp. 818–828.

Kirsch, D., Goldfarb, B., and Gera, A. 2009. "Form or substance: the role of business plans in venture capital decision making," *Strategic Management Journal* (30:5), pp. 487–515.

Landström, H. 1998. "Informal investors as entrepreneurs: Decision-making criteria used by informal investors in their assessment of new investment proposals," *Technovation* (18:5), pp. 321–333.

Larrick, R. P., Mannes, A. E., Soll, J. B., and Krueger, J. I. 2011. "The social psychology of the wisdom of crowds," *Social Psychology and Decision Making*, pp. 227–242.

Li, H. 2001. "How does new venture strategy matter in the environment–performance relationship?" *The Journal of High Technology Management Research* (12:2), pp. 183–204.

Magder, L. S., and Hughes, J. P. 1997. "Logistic regression when the outcome is measured with uncertainty," *American Journal of Epidemiology* (146:2), pp. 195–203.

Magnusson, P. R., Wästlund, E., and Netz, J. 2016. "Exploring users' appropriateness as a proxy for experts when screening new product/service ideas," *Journal of Product Innovation Management (33:1),* pp. 4–18.

Malone, T. W., Laubacher, R., and Dellarocas, C. 2009. "Harnessing crowds: Mapping the genome of collective intelligence".

Mannes, A. E., Soll, J. B., and Larrick, R. P. 2014. "The wisdom of select crowds," *Journal of Personality and Social Psychology* (107:2), p. 276.

Marino, K. E., and Noble, A. F. de 1997. "Growth and early returns in technology-based manufacturing ventures," *The Journal of High Technology Management Research* (8:2), pp. 225–242.

Mason, C., and Stark, M. 2004. "What do investors look for in a business plan? A comparison of the investment criteria of bankers, venture capitalists and business angels," *International Small Business Journal* (22:3), pp. 227–248.

Matthews, B. W. 1975. "Comparison of the predicted and observed secondary structure of T4 phage lysozyme," *Biochimica et Biophysica Acta (BBA)-Protein Structure* (405:2), pp. 442–451.

Maxwell, A. L., Jeffrey, S. A., and Lévesque, M. 2011. "Business angel early stage decision making," *Journal of Business Venturing* (26:2), pp. 212–225.

Meehl, P. E. 1955. *Clinical vs. Statistical Prediction: A Theoretical Analysis and a Review of the Evidence*: Sage Publications.

Mellers, B., Stone, E., Atanasov, P., Rohrbaugh, N., Metz, S. E., Ungar, L., Bishop, M. M., Horowitz, M., Merkle, E., and Tetlock, P. 2015. "The psychology of intelligence analysis: Drivers of prediction accuracy in world politics," *Journal of Experimental Psychology: Applied* (21:1), pp. 1–14.

Morris, R. D. 1987. "Signalling, agency theory and accounting policy choice," *Accounting and Business Research* (18:69), pp. 47–56.

Nagar, Y., and Malone, T. 2011. "Making business predictions by combining human and machine intelligence in prediction markets," *Proceedings of the International Conference on Information Systems 2011.*

Nambisan, S. 2016. "Digital Entrepreneurship: Toward a Digital Technology Perspective of Entrepreneurship," *Entrepreneurship Theory and Practice.*

Nickerson, R. C., Varshney, U., and Muntermann, J. 2013. "A method for taxonomy development and its application in information systems," *European Journal of Information Systems* (22:3), pp. 336–359.

Olson, D. L., Delen, D., and Meng, Y. 2012. "Comparative analysis of data mining methods for bankruptcy prediction," *Decision Support Systems* (52:2), pp. 464–473.

Önkal, D., Gönül, M. S., Goodwin, P., Thomson, M., and Öz, E. 2017. "Evaluating expert advice in forecasting: Users' reactions to presumed vs. experienced credibility," *International Journal of Forecasting* (33:1), pp. 280–297.

Petersen, M. A. 2004. "Information: Hard and soft"

Quinn, A. J., and Bederson, B. B. 2011. *Human computation: a survey and taxonomy of a growing field,* ACM.

Riedl, C., Blohm, I., Leimeister, J. M., and Krcmar, H. 2013. "The effect of rating scales on decision quality and user attitudes in online innovation communities," *International Journal of Electronic Commerce* (17:3), pp. 7–36.

Shepherd, D. A., and Zacharakis, A. 1999. "Conjoint analysis: A new methodological approach for researching the decision policies of venture capitalists," *Venture Capital: An International Journal of Entrepreneurial Finance* (1:3), pp. 197–217.

Shmueli, G., and Koppius, O. R. 2011. "Predictive analytics in information systems research," *MIS Quarterly*, pp. 553–572.

Silver, N. 2012. *The signal and the noise: why so many predictions fail--but some don't:* Penguin.

Simon, H. A. 1955. "A behavioral model of rational choice," *The Quarterly Journal of Economics* (69:1), pp. 99–118.

Simon, H. A. 1982. *Models of bounded rationality: Empirically grounded economic reason:* MIT Press.

Song, M., Podoynitsyna, K., van der Bij, H., and Im Halman, J. 2008. "Success factors in new ventures: A meta-analysis," *Journal of Product Innovation Management* (25:1), pp. 7–27.

Spence, M. 1974. "Competitive and optimal responses to signals: An analysis of efficiency and distribution," *Journal of Economic theory* (7:3), pp. 296–332.

Spiegel, O., Abbassi, P., Zylka, M. P., Schlagwein, D., Fischbach, K., and Schoder, D. 2015. "Business model development, founders' social capital and the success of early stage internet start-ups: a mixed-method study," *Information Systems Journal* (26:5), pp. 421–449.

Sudek, R. 2006. "Angel investment criteria," *Journal of Small Business Strategy* (17:2), p. 89.

Todd, P., and Benbasat, I. 1999. "Evaluating the impact of DSS, cognitive effort, and incentives on strategy selection," *Information Systems Research* (10:4), pp. 356–374.

Valle, M. A., Varas, S., and Ruz, G. A. 2012. "Job performance prediction in a call center using a naive Bayes classifier," *Expert Systems with Applications* (39:11), pp. 9939–9945.

van Bruggen, G. H., Spann, M., Lilien, G. L., and Skiera, B. 2010. "Prediction markets as institutional forecasting support systems," *Decision Support Systems* (49:4), pp. 404–416.

Vapnik, V. N. 1998. *Statistical Learning Theory.* New York: Wiley.

Webster, J., and Watson, R. T. 2002. "Analyzing the past to prepare for the future: Writing a literature review," *MIS Quarterly* (26:2), pp. xiv- xxiii.

Woolley, A. W., Chabris, C. F., Pentland, A., Hashmi, N., and Malone, T. W. 2010. "Evidence for a collective intelligence factor in the performance of human groups," *Science* (330:6004), pp. 686–688.

Yuan, H., Lau, R. Y. K., and Xu, W. 2016. "The determinants of crowdfunding success: A semantic text analytics approach," *Decision Support Systems* (91), pp. 67–76.

Zahra, S. A., and Bogner, W. C. 2000. "Technology strategy and software new ventures' performance: Exploring the moderating effect of the competitive environment," *Journal of Business Venturing* (15:2), pp. 135–173.

Zahra, S. A., Matherne, B. P., and Carleton, J. M. 2003. "Technological resource leveraging and the internationalization of new ventures," *Journal of International Entrepreneurship* (1:2), pp. 163–186.

*and collective intelligence to demonstrate its utility for predictions under extreme uncertainty.*